\documentclass[letterpaper, 10 pt, conference]{ieeeconf}  
\usepackage{amsmath,amsfonts}
\usepackage{algorithmicx}
\usepackage{gensymb}
\usepackage{array}
\usepackage[caption=false,font=normalsize,labelfont=sf,textfont=sf]{subfig}
\usepackage{textcomp}
\usepackage{stfloats}
\usepackage{url}
\usepackage{verbatim}
\usepackage{graphicx}
\usepackage{color}
\usepackage{multirow}
\usepackage{hyperref}
\usepackage{algorithm}
\usepackage{algpseudocode}
\usepackage{amssymb}

\usepackage{subfloat}

\IEEEoverridecommandlockouts  

\title{\LARGE \bf
Multi-Agent Generative Adversarial Interactive Self-Imitation Learning for 
AUV Formation Control and Obstacle Avoidance 
}

\author{Zheng Fang$^{\dagger*}$, Tianhao Chen$^{\dagger*}$, Dong Jiang$^{\dagger}$, Zheng Zhang$^{\dagger}$ and Guangliang Li$^{\dagger**}$
\thanks{$^{\dagger}$Zheng Fang, Tianhao Chen, Dong Jiang, Zheng Zhang and Guangliang Li are with the School of Information Science and Engineering, 
Ocean University of China, 
Qingdao, China. 
		{\tt\small 
		\{guangliangli\}@ouc.edu.cn}}
\thanks{$^{*}$ Contributing equally}  
\thanks{$^{**}$ Corresponding author}
}

\begin{document}
\maketitle
\thispagestyle{empty}
\pagestyle{empty}

\begin{abstract}


Multiple autonomous underwater vehicles (multi-AUV) can cooperatively accomplish tasks that a single AUV cannot complete. 
Recently, multi-agent reinforcement learning has been introduced to control of multi-AUV. However, designing efficient reward functions for various tasks of multi-AUV control is difficult or even impractical. Multi-agent generative adversarial imitation learning (MAGAIL) allows multi-AUV to learn from expert demonstration instead of pre-defined reward functions, but suffers from the deficiency of requiring optimal demonstrations and not surpassing provided expert demonstrations. This paper builds upon the MAGAIL algorithm by proposing multi-agent generative adversarial interactive self-imitation learning (MAGAISIL), which can facilitate AUVs to learn policies by gradually replacing the provided sub-optimal demonstrations with self-generated good trajectories
selected by a human trainer. Our experimental results in a multi-AUV
formation control and obstacle avoidance task on the Gazebo platform with AUV simulator of our lab show that AUVs trained via MAGAISIL can surpass the provided sub-optimal expert demonstrations and reach a performance close to or even better than MAGAIL with optimal demonstrations. Further results indicate that AUVs' policies trained via MAGAISIL can adapt to complex and different tasks as well as MAGAIL learning from optimal demonstrations.

\end{abstract}

\section{INTRODUCTION}



Autonomous underwater vehicle (AUV) plays an important role in underwater tasks of exploring marine resources and scientific research due to its flexibility \cite{paull2013auv,cheng2021path}. It can replace humans to perform dangerous underwater tasks 
such as survey of ocean topography and landforms, inspection, maintenance and repair of submarine oil pipelines.
Reinforcement learning (RL) was introduced and applied to improve the autonomy and intelligence of AUV control \cite{kober2013reinforcement,carreras2005behavior,el2008policy,el2013two,yu2017deep,zhang2020deep,grando2021deep,hadi2022deep,zhang2022auv}. Through interactions with underwater environment, AUV with RL can learn a control policy adapting to the changes and uncertainty \cite{yu2017deep,carlucho2018adaptive}. However, as the limited detection range and energy storage of a single AUV, it is necessary to use multi-AUV to cooperatively complete underwater tasks that single AUV cannot perform with the increasing complexity of underwater missions. Multiple AUVs can accomplish underwater detection, target search, object recognition etc., in a collaborative way, which can improve the efficiency of task execution and reduce the time and energy cost. Multi-agent reinforcement learning (MARL) has been introduced to improve multi-AUV control in uncertain marine environments \cite{fang2022autonomous,wang2022multi,lin2022smart}. However, it is difficult to design efficient reward functions for various tasks, especially those complex and high-dimensional ones where most robots like AUVs will be operated in. Moreover, the difficulty of designing reward functions for MARL increases with the number of agents and complexity of their relationships \cite{gu2004multiagent,bucsoniu2010multi}.

Imitation learning was proposed and successfully applied to robot control \cite{argall2009survey,ravichandar2020recent} since it is much easier to provide demonstrations on performing a task than to design a reward function. There are mainly two kinds of imitation learning: one is behavior cloning (BC) and the other is inverse reinforcement learning (inverse RL). BC learns a mapping from an agent's states to optimal actions via supervised learning \cite{ross2010efficient}, but requires a large amount of data and cannot generalize to unseen situations and adapt to different tasks effectively. Inverse reinforcement learning agents learn control policies with extracted cost functions from expert demonstrations via reinforcement learning \cite{ng2000algorithms} and can effectively generalize to unseen states \cite{ho2016model}. However, many inverse RL algorithms need a model to solve a sequence of planning or reinforcement learning problem in an inner loop and the performance might decrease if the planning or RL problem is not optimally solved \cite{ho2016generative}, which prevents applying inverse RL for robot control to large and complex tasks. Ho et al. solved this problem by proposing a general model-free imitation learning method --- generative adversarial imitation learning (GAIL) \cite{ho2016generative}, which allows robots to directly learn policies from expert demonstrations in large and complex environments. Higaki et al. applied GAIL to realize ship's automatic collision avoidance by mimicking human expert performance \cite{higaki2023human}. Jiang et al. \cite{jiang2022generative} implemented GAIL in AUV path following tasks and further proposed a generative adversarial interactive imitation learning (GA2IL) method combining GAIL with interactive RL \cite{li2019human,christiano2017deep} to improve AUV's performance and stability in path following. 

GAIL was extended to a multi-agent setting by proposing multi-agent generative adversarial imitation learning (MAGAIL) \cite{song2018multi}. Fang et al. \cite{fang2022autonomous} successfully applied MAGAIL to a multi-AUV formation control task with a decentralized training and execution framework. However, MAGAIL shares the limitation with GAIL and other imitation learning methods that they assume the optimality of expert demonstrations and can seldom surpass the performance of demonstrations if the provided demonstrations are not optimal. On the other hand, Guo et al. assumed that optimal demonstrations are not available and agents should imitate ``relatively better trajectories" generated by the agent. They proposed generative adversarial self-imitation learning (GASIL) \cite{guo2018generative} by imitating agent's past good trajectories measured via pre-defined reward functions, which violates the initial idea of the GAIL framework learning from solely demonstrations and avoiding pre-defined reward functions. 

In this paper, we proposed multi-agent generative adversarial interactive self-imitation learning (MAGAISIL) by improving MAGAIL via replacing the provided expert sub-optimal demonstrations with agent generated good trajectories. However, different from GASIL, MAGAISIL allows a human trainer to evaluate whether the agent generated trajectories are better than the provided expert demonstrations instead of using pre-defined reward functions. 
We implemented and tested our MAGAISIL method in a multi-AUV formation control and obstacle avoidance task on the Gazebo platform with AUV simulator of our lab, and compared to the MAGAIL method. Our results show that, even provided with sub-optimal expert demonstrations, AUVs trained with our MAGAISIL method can learn to reach a performance close to or even better than those trained via
MAGAIL with optimal demonstrations via gradually replacing the sub-optimal demonstrations with self-generated good trajectories selected by a human trainer. In addition, further results indicate that the control policies of AUVs trained via MAGAISIL can adapt to complex and different tasks even with added obstacles or changed angles of wall as well as MAGAIL learning from optimal demonstrations.

\section{Background}

\subsection{Multi-Agent Reinforcement Learning}

In single-agent reinforcement learning \cite{sutton2018reinforcement,arulkumaran2017deep,kober2013reinforcement}, the objective of an agent is to learn a policy maximizing cumulative returns via interacting with the environment. For example, at time step $t$, an agent selects an action $a_t$ with its policy based on the detected current state $s_t$. Then the agent will transition to a next state and receive a reward $r_t$ from the environment. The goal of the agent is to learn a policy $\pi$ which maximizes the discounted accumulated return as follow:
\begin{equation}
	V_\pi (s) = \mathbb{E} \bigg[ \sum_{h \geq 0} \gamma^h r_t \big| a_t \sim \pi(a|s_t) \bigg],
\end{equation}
where $\gamma$ is the discount factor determining the value of future rewards over immediate ones, $V_\pi (s)$ is the value of state $s$ following policy $\pi$ thereafter.

In multi-agent reinforcement learning (MARL), there are multiple agents interacting with the environment \cite{busoniu2008comprehensive,qie2019joint}. Each agent $i$ has its own policy $\pi_i$ that can be used to select an action $a_{i,t}$ based on its observed state $s_t$ at current time step $t$. Then the agent transitions to a next state and will receive a reward $r_{i,t}$. Similar to single-agent reinforcement learning, the goal for each agent is to learn a policy maximizing its discounted accumulated return. However, different from single-agent reinforcement learning, in MARL, the policy $\pi_i$ of agent $i$ is affected by other agents' policies. The most common concept to solve this problem is Nash Equilibrium (NE). In NE, agent $i$ will not try to change its policy $\pi_i$ if other agents do not change their policies, because its discounted accumulated return cannot continue to increase. That is to say, if all agents reach the equilibrium state, 
each learns a steady optimal policy. 

In MARL, the relationship between agents can be divided into three settings based on the relationship between reward functions of agents: cooperative, competitive and a mixed setting \cite{zhang2021multi}. In a fully cooperative setting, all agents perform the same task and share a same reward function. The relationship between reward functions of agents is zero-sum in a competitive setting. In other words, agents maximize their cumulative rewards by preventing each other from completing its task. In a mixed setting, each agent has its own task and reward function, which can be cooperative or competitive to other agents. In our experiments, the relationship between leader and follower AUV is in a mixed setting 
since they perform different tasks.

\subsection{Generative Adversarial Imitation Learning}


Generative adversarial imitation learning (GAIL) \cite{ho2016generative} allows an agent to learn directly from expert demonstrations consisting of state-action pairs, avoiding to pre-define reward functions for various tasks. 
A GAIL agent trained a discriminator $D: S \times A \rightarrow (0,1)$ to distinguish expert state-action pairs $(s, a) \sim \tau_E$ from agent state-action pairs $(s, a) \sim \tau_{agent}$, 
and a generator (i.e., policy $\pi$) to ``fool" the discriminator by generating 
state-action pairs $(s, a) \sim \tau_{agent}$ as close as possible to expert state-action pairs $(s, a) \sim \tau_E$ by maximizing $\mathbb{E}_\pi[log(D(s,a))]$. 
The agent generates its trajectory $\tau_{agent}$ by interacting with the environment with its current policy $\pi$.
That is to say, a GAIL agent learns a policy directly by generating a distribution of the agent's state-action pairs as close as possible to the distribution of state-action pairs from the expert demonstrations.
In summary, the GAIL algorithm can be summarized as finding a saddle point ($\pi$, $D$): 
\begin{equation}
- \lambda H(\pi) +\mathbb{E}_\pi[log(D(s,a))] + \mathbb{E}_{\pi_E}[log(1-D(s,a))],
\label{con:gail}
\end{equation}
where $H(\pi)\triangleq\mathbb{E}_\pi[-log \pi(a\mid s)]$, is the $\gamma$-discounted causal entropy \cite{bloem2014infinite} of the policy $\pi$, $\lambda$ is the weight of entropy $H(\pi)$. 
Song et al. extended GAIL to a multi-agent setting by proposing multi-agent generative adversarial imitation learning (MAGAIL) \cite{song2018multi}. Guo et al. assumed that optimal demonstrations are not always available to an agent in RL, and proposed generative adversarial self-imitation learning (GASIL) to imitate agent's past good trajectories by measuring them via pre-defined reward functions \cite{guo2018generative}.

\section{Methodology}

\begin{figure}[!t]
\centering
\vspace{3mm}
\includegraphics[width=3.3in]{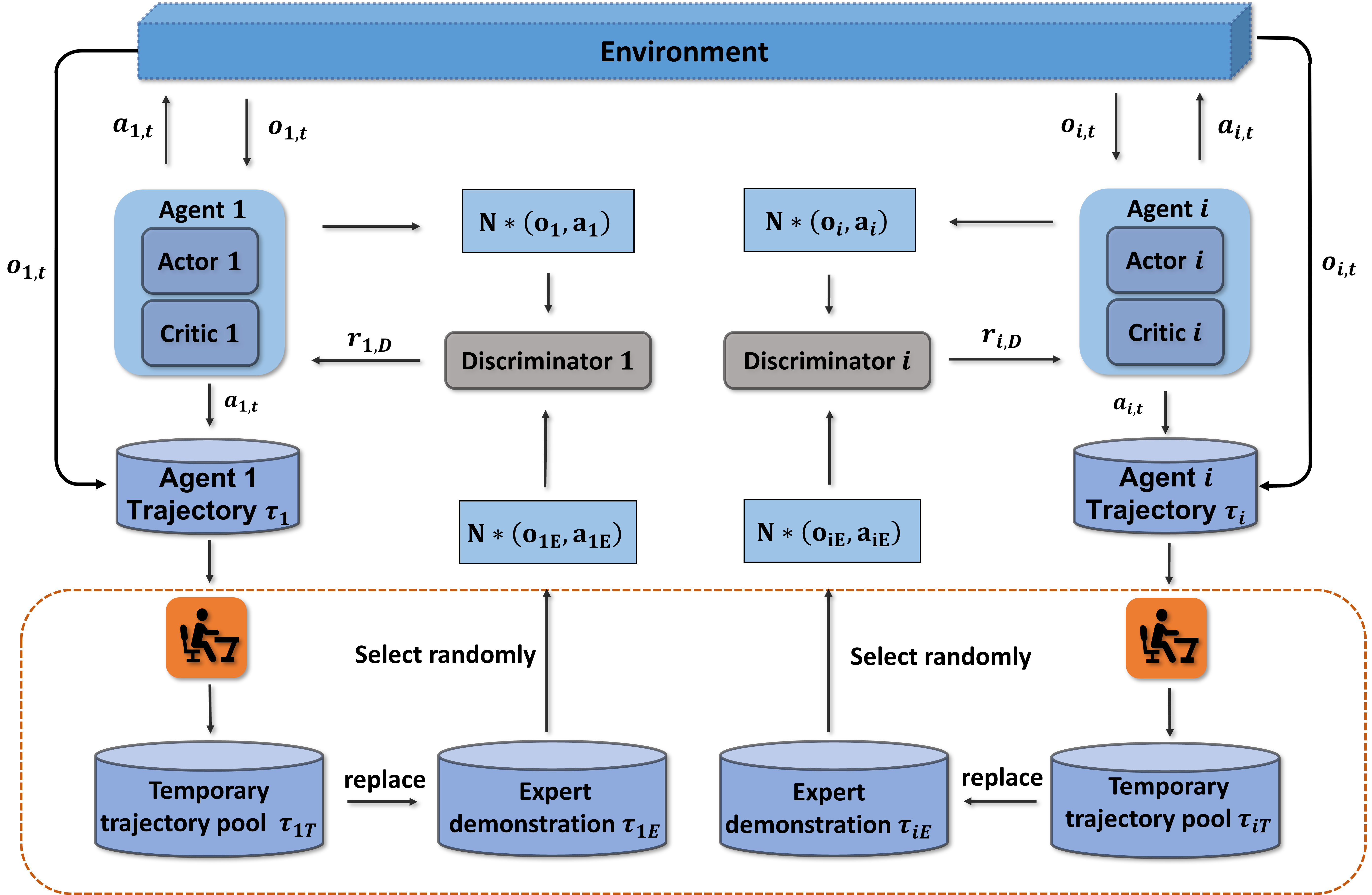}
\caption{Illustration of the mechanism for our multi-agent generative adversarial interactive self-imitation learning (MAGAISIL) method.} 
\label{magaisil}
\vspace{-6mm}
\end{figure}


Previous work extending GAIL to multi-agent learning by proposing MAGAIL \cite{song2018multi}, which allows multiple agents to learn from provided expert demonstrations. However, MAGAIL shares the limitation with GAIL and other imitation learning methods that they can seldom surpass the performance of demonstrations. On the other hand, generative adversarial self-imitation learning (GASIL) \cite{guo2018generative} aims to imitate agent's past good trajectories by measuring them via pre-defined reward functions, but violates the initial idea of the GAIL framework to allow learning from demonstrations and avoid pre-defining reward functions. In this paper, we proposed multi-agent generative adversarial interactive self-imitation learning (MAGAISIL) by improving MAGAIL via replacing the provided expert sub-optimal demonstrations with agent generated good trajectories. However, different from GASIL, MAGAISIL allows a human trainer to evaluate whether the agent generated trajectories are better than the provided expert demonstrations instead of using pre-defined reward functions. Therefore, we expect and hypothesize that our MAGAISIL method allow agents to learn from solely sub-optimal expert demonstrations without pre-defining reward functions and can obtain a much better performance than the provided expert demonstrations, resolving the limitation of MAGAIL that it can seldom surpass the performance of demonstrations. Fig. \ref{magaisil} illustrates the mechanism of our proposed MAGAISIL method.



As shown in Fig. \ref{magaisil}, MAGAISIL will take provided sub-optimal expert demonstrations as input. Then, each agent will learn an Actor (i.e. control policy) and a Critic (i.e. value function) via independent proximal policy optimization (IPPO) \cite{schroeder2020independent}. In addition, a discriminator will be trained to distinguish the state-action pairs of agent's generated trajectories from provided expert demonstrations. Specifically, during training, at time step $t$, Agent $i$ will obtain local observation ${o}_{i, t}$, and select an action $a_{i, t}$ with its current policy $\pi_{\theta_i}$: $\quad a_{i, t} \sim \pi_{\theta_i}\left(a_i \mid o_{i, t}\right)$. Then, it will transition to a new state upon performing the selected action. The Agent $i$ will repeat the cycle of selecting action and obtaining observation until the end of an episode. The received state-action pairs during one episode by interacting with the environment compose the trajectory $\tau_i$ of Agent $i$. 
$N$ state-action pairs $({o}_{i}, {a}_{i})$ from the agent's trajectory $\tau_i$ will be selected and $N$ state-action pairs $({o}_{iE}, {a}_{iE})$ from the provided expert trajectory $\tau_{i E}$ will be selected 
and used to train the discriminator $D_{\omega_i}$ via ADAM \cite{zhang2018improved} with the loss function as:
\begin{equation}
\label{MAGAIL loss}
\mathbb{E}_{\tau_i}\left[\log \left(D_{\omega_i}(o, a)\right)\right]+\mathbb{E}_{\tau_{i E}}\left[\log \left(1-D_{\omega_i}(o, a)\right)\right].
\end{equation}
The updated discriminator $D_{\omega_i}$ will be used to provide rewards $r_{i,D}$ for updating the Actor and Critic of Agent $i$:
\begin{equation}
r_{i, D}=-\log \left(1-D_{\omega_i}\left(o, a\right)\right).
\end{equation}

In addition, at the end of an episode, the trajectory $\tau_i$ of Agent $i$ will be visualized and shown to a human trainer, who can compare with the provided expert demonstrations according to her knowledge and experience. If the human trainer thinks the agent's generated trajectory is better than the expert demonstrations, the trajectory $\tau_i$ of Agent $i$ will be stored in the temporary trajectory pool $\tau_{i_T}$. Otherwise, the trajectory $\tau_i$ of Agent $i$ will be disregarded. We set a limit to the number of state-action pairs in the pool $\tau_{i_T}$ and when it is full, all stored trajectories in $\tau_{i_T}$ are used to replace the current expert demonstrations $\tau_{i_E}$. At the same time, $\tau_{i_T}$ will be cleared to store new trajectories in the following training process. 

\section{Simulation Setup}

\begin{figure}[!t]
\centering
\vspace{3mm}
\includegraphics[width=3in]{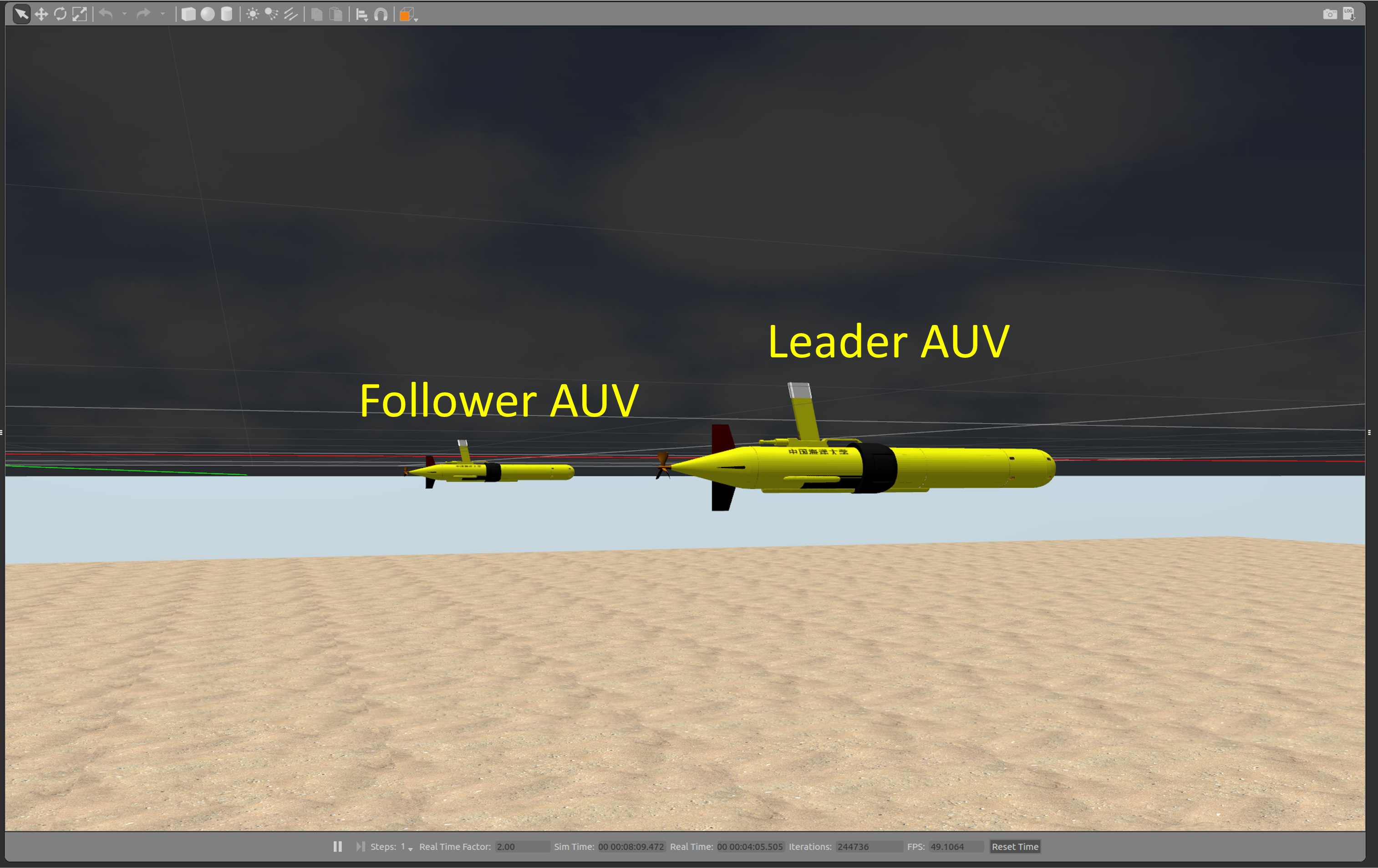}
\caption{Screenshot of the Gazebo simulation platform with 
a leader AUV and a follower AUV in the simulated underwater environment.} 
\vspace{-6mm}
\label{simulator}
\end{figure}

We evaluated our method by conducting experiments in three formation control and obstacle avoidance of multi-AUV tasks on the Gazebo simulation platform. The simulator is extended from Unmanned Underwater Vehicle Simulator \cite{manhaes2016uuv} with the model of Sailfish 210 developed in our lab. Fig. \ref{simulator} shows the screenshot of Gazebo simulation platform with a leader AUV and a follower AUV in the simulated underwater environment. 

\begin{figure*}[!h]
	\centering
        \setlength{\belowcaptionskip}{-10mm}
	\subfloat[Task I]{
		\includegraphics[width=5.78cm]{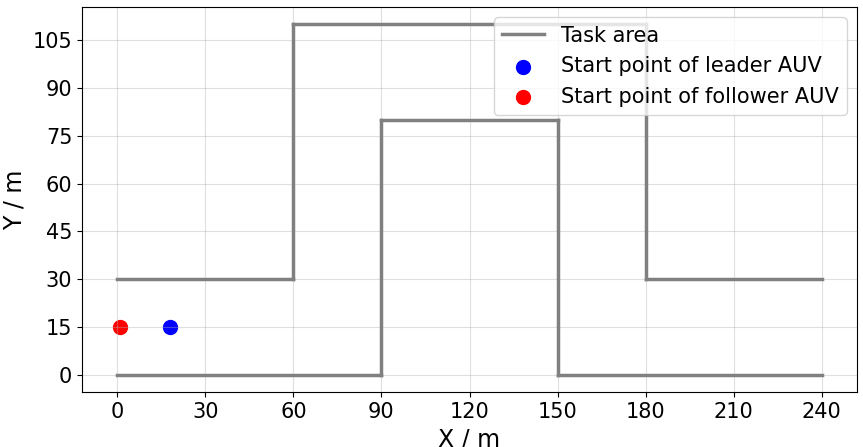}
	}
	\subfloat[Task II]{
		\includegraphics[width=5.78cm]{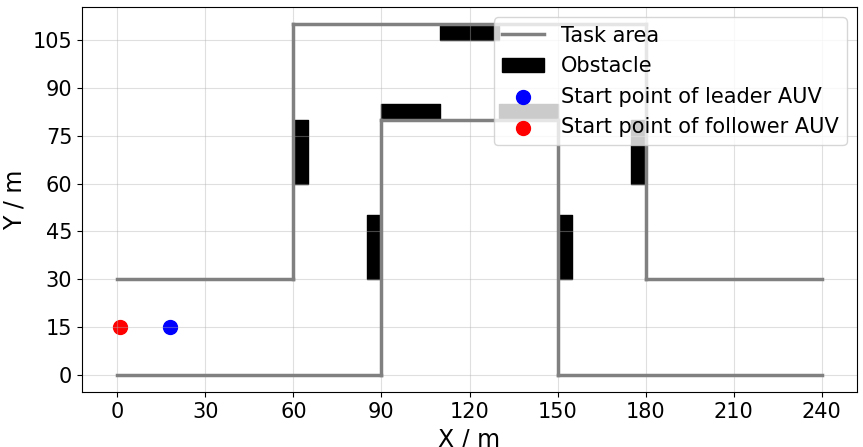}
	}
	\subfloat[Task III]{
		\includegraphics[width=5.78cm]{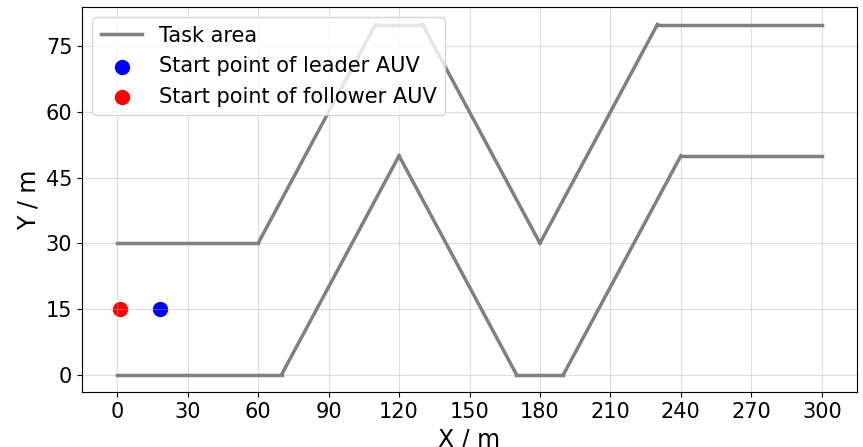}
	}
        \vspace{-2mm}
        \caption{\textcolor{black}{Illustration of the configuration of Task I, II and III in the underwater environment for formation
control and obstacle avoidance.} }
        \label{task area}
        \vspace{-4mm}

\end{figure*}

\begin{figure}[!t]
\centering
\vspace{-3mm}
\includegraphics[width=2.8in]{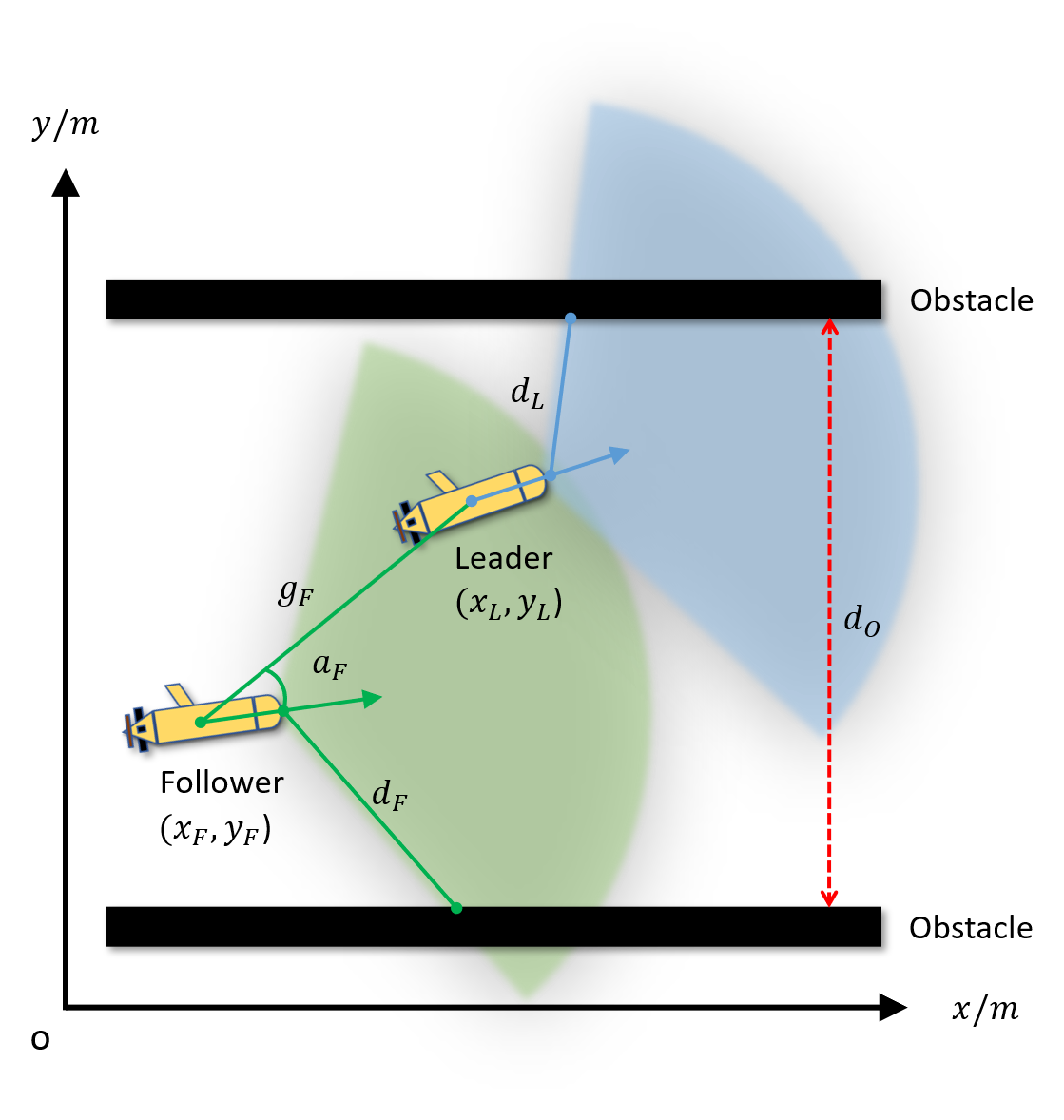}
\vspace{-6mm}
\caption{The state representation of the leader and follower AUV in the tasks.} 
\vspace{-4mm}
\label{state}
\end{figure}

\subsection{Simulation Tasks}
\label{tasks}

We set up three formation control and obstacle avoidance tasks in our experiments as shown in Fig. \ref{task area}. Fig. \ref{state} shows the observed state information of a leader AUV and a follower AUV in the tasks. In Task I, the objective of the task is to allow the leader AUV to go through a square pipe and the follower AUV to follow the leader AUV at a distance of $18$ meters, while keeping a safe distance from the wall of both sides in the pipe. The distance between walls of two sides in the pipe is 30 meters. The leader AUV and follower AUV will start from the position $(18, 0)$ and $(1, 0)$, respectively, and the ending position is at $(240, 0)$. The task is terminated and a new episode will start in the following situations:

\begin{enumerate}
\item{Leader AUV or follower AUV is too close to obstacles, i.e.,$\left|d_L\right| \leq 2 \text { or }\left|d_F\right| \leq 2$ }
\item{The distance between two AUVs is too close or far, i.e.,$\left|g_F\right|<3 \text { or }\left|g_F\right|>33$}
\item{The heading deviation of follower AUV is too large, i.e.,$\left|a_F\right| \geq \frac{\pi}{3}$}.
\end{enumerate}

To test the adaptability of our method in complex tasks, we added rectangular obstacles with a length of 20 meters and width of 5 meters along the walls on both sides of the pipe in Task II, and changed the angles of walls and extended the pipe to be 300 meters long in Task III.

\subsection{State Representation and Action Space}

In the tasks, the leader-follower method was adopted and a leader AUV and a follower AUV were considered for simplification, which can be easily extended to complex tasks with more AUVs, as shown in Fig. \ref{simulator}. A decentralized training and execution framework was used as in \cite{fang2022autonomous}. Fig. \ref{state} shows the state representation of AUVs in the task. 
As shown in Fig. \ref{state}, the black squares represent the walls of pipe or obstacles, $d_o$ denotes the distance between walls of the pipe, which is set to be 30 meters, $\left(x_L, y_L\right)$ is the coordinate of the leader AUV's current position, and $\left(x_F, y_F\right)$ is the coordinate of the follower AUV's current position. Both AUVs need to detect and avoid collision with the wall and/or obstacles with sonar sensor and go through the pipe as soon as possilbe. The detection angle range of the sonar sensor is set to be $\left[-\frac{\pi}{3}, \frac{\pi}{3}\right]$ which will be divided into 6 sectors, as shown by the shaded blue and green area in Fig. \ref{state}. The sonar sensor has 600 beams which are equally distributed in the 6 sectors. The detection distance is 33 meters at most. AUV will take the shortest distance detected by all the sectors as the distance to the obstacle. The detected shortest distances by the leader AUV and follower AUV are denoted as $d_L$ and $d_F$, respectively. \textcolor{black}{The leader AUV's state is represented as $o_L=\left\{d_{L 1}, d_{L 2}, d_{L 3}, d_{L 4}, d_{L 5}, d_{L 6}\right\}$} and the follower AUV's state is represented as $o_F=\left\{g_F, a_F, d_{F 1}, d_{F 2}, d_{F 3}, d_{F 4}, d_{F 5}, d_{F 6}\right\}$. Here, $d_{L i}$ and $d_{F i}$ are the detected distance by each sector of the sonar sensor on leader AUV and follower AUV, respectively, $i=1,2,...,6$. $g_F$ is the directrix distance between two AUVs, which is computed as $g_F=\sqrt{\left(y_F-y_L\right)^2+\left(x_F-x_L\right)^2}$, and $a_F$ is the heading deviation of follower AUV and computed as $a_F=a_{F H}-\operatorname{atan} 2\left(\frac{y_F-y_L}{x_F-x_L}\right)$, 
\textcolor{black}{where} $a_{FH}$ denotes the current heading of the follower AUV. 

\begin{table}[!htbp]
	\caption{
 Action setup for AUVs to perform by setting the angles of four rudders \textcolor{black}{(upper, right, lower and left)}} 
	\label{actions}
	\centering
	\begin{tabular*}{6cm}{ccccc}
		\hline
		Action & Upper & Right & Lower & Left \\
		\hline
		turn left 1 & $-14^{\circ}$ & $0$ & $14^{\circ}$ & $0$ \\
		turn left 2 & $-20^{\circ}$ & $0$ & $20^{\circ}$ & $0$ \\
		go straight & $0$ & $0$ & $0$ & $0$ \\
		turn right 1 & $14^{\circ}$ & $0$ & $-14^{\circ}$ & $0$ \\
		turn right 2 & $20^{\circ}$ & $0$ & $-20^{\circ}$ & $0$ \\
		\hline
	\end{tabular*}
    \vspace{-1.em}
\end{table}


We set five discrete actions for both AUVs, which can be performed by setting the thruster speed and angles of four rudders, including two actions for turning left, two for turning right and one for going straight, as shown in Table \ref{actions}. The upper and lower rudders are set to control the horizontal direction of AUV. The left and right rudders are used to control AUV to float and dive in the underwater environment, which are set to 0 as the tasks are in a 2D space. The thruster speed of AUV is set to 300 $r/s$.


\begin{figure*}
	\centering
	\subfloat[Learning curve for the leader AUV]{
        \begin{minipage}{0.45\linewidth}
			\centering
		   \includegraphics[width=7cm]{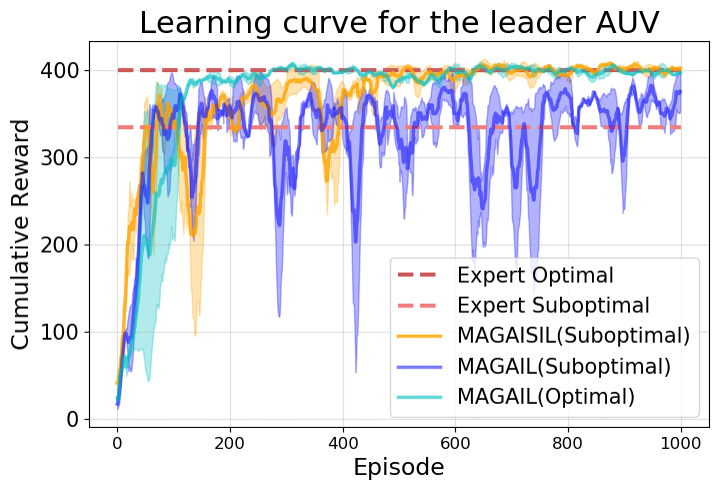}
		\end{minipage}
	}
	\subfloat[Learning curve for the follower AUV]{
	   \begin{minipage}{0.45\linewidth}
			\centering
		    \includegraphics[width=7cm]{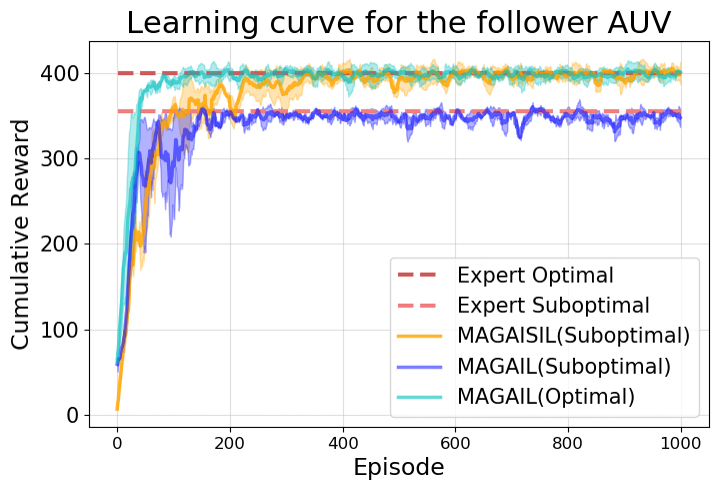}
	   \end{minipage}
	}
	\vspace{-1mm}
	\caption{Cumulative rewards received by the leader AUV and follower AUV trained via MAGAISIL with suboptimal demonstrations (MAGAILSIL Optimal), MAGAIL with sub-optimal (MAGAIL Suboptimal) and optimal (MAGAIL Optimal) demonstrations in Task I. The shaded area is the 0.95 confidence interval and the bold line is the mean performance over three experimental trials. Two red lines show the performance of expert optimal and sub-optimal demonstrations.}
	\label{learningcurve}
	\vspace{-6mm}
\end{figure*}

\begin{figure*}[!h]
	\centering
        \setlength{\belowcaptionskip}{-10mm}
	\subfloat[Task I]{
		\includegraphics[width=5.78cm]{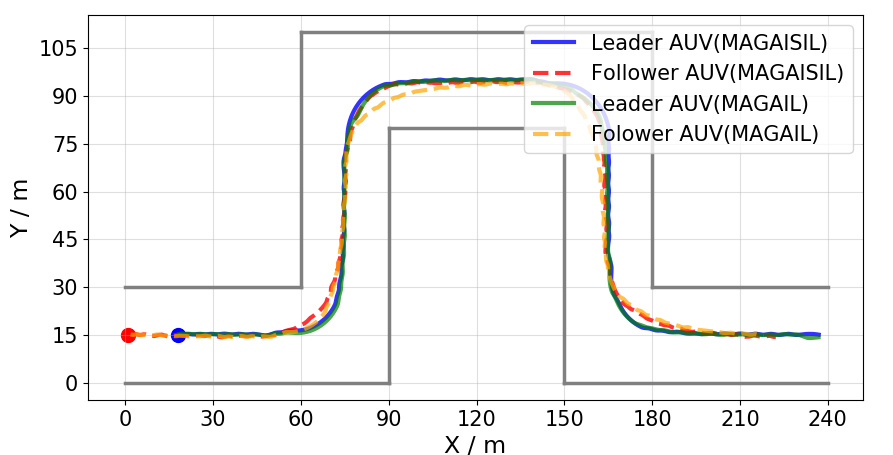}
	}
	\subfloat[Task II]{
		\includegraphics[width=5.78cm]{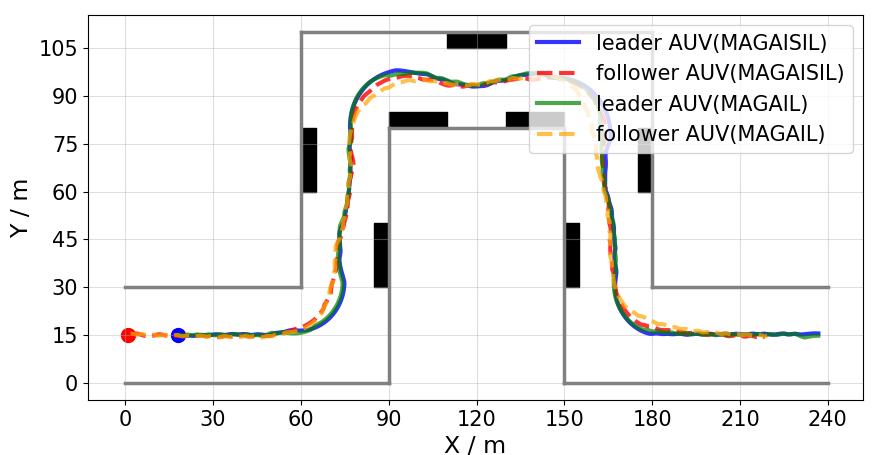}
	}
	\subfloat[Task III]{
		\includegraphics[width=5.78cm]{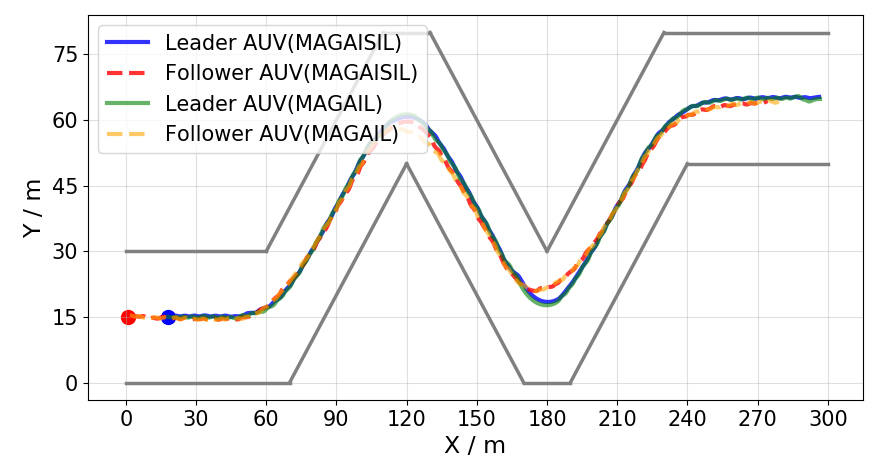}
	}
        \vspace{-3mm}
        \caption{Tested trajectories of the leader AUV and follower AUV in Task I, II and III using final control policies trained in Task I via MAGAISIL with sub-optimal expert demonstrations and MAGAIL with optimal expert demonstrations.}
        \label{trajectory}
        \vspace{-6mm}

\end{figure*}


\subsection{Evaluation Metrics}
\label{em}

Due to the subjectivity of rewards from the learned discriminator with expert demonstrations, we defined reward functions for both leader and follower AUVs to evaluate our proposed method. The defined reward functions for both leader AUV and follower AUV are never used for learning, but only used for testing the learned policies from demonstrated trajectories with MAGAIL and our method MGAISIL. 
The reward function for the leader AUV is defined as:
\begin{equation}
\label{section-reward}
r_L=1-\frac{\left|d_L-17.3\right|}{8.65},
\end{equation}
where 17.3 meters is a safe distance for AUV derived based on the detection angle range of the sonar sensor and the distance between walls of the pipe, which can keep AUV in the middle of the pipe.
The reward function for the follower AUV is defined as:
\begin{equation}
\label{leader reward}
r_F=0.5 * r_F^c+0.5 * r_F^a,  
\end{equation}
where $r_F^c$ represents the rewards for tracking the leader AUV, and is defined as $r_F^c=-\frac{\left|a_F\right| * 3}{\pi}+\left|1-\frac{\left|g_F-18\right|}{15}\right|$ based on the distance $g_L$ between two AUVs and the heading deviation $a_L$. Similar to the leader AUV, $r_F^a=1-\frac{\left|d_F-17.3\right|}{8.65}$ is defined based on the distance from the wall of pipe $d_F$ to keep the follower AUV in the middle of the pipe. 


In our experiments, we trained both leader and follower AUV with our MAGAISIL method with sub-optimal expert demonstrations. In addition, MAGAIL trained with both optimal and sub-optimal expert demonstrations were also used as comparisons.
\textcolor{black}{We set $N = 256$, i.e., each time 256 state-action pairs  are randomly selected from trajectory generated by each AUV and provided expert demonstrations respectively to update the discriminator. During one episode, the discriminator is updated $3$ times and the generator is updated $9$ times for each AUV.}
The discount factor is set to be $\gamma=0.99$, $\lambda=1.0$ and the clipping factor $\epsilon=0.09$ in the IPPO algorithm. \textcolor{black}{The maximum number of state-action pairs in the temporary trajectory pool is set to be 2000.} 

\section{Results and Discussion}


This section presents and analyzes our experimental results by comparing the policy performance trained with our MAGAISIL learning from sub-optimal expert demonstrations to MAGAIL learning from both optimal and sub-optimal expert demonstrations in Task I as described in Section \ref{tasks}. In addition, to evaluate the adaptability of our MAGAISIL method, we tested the trained policies of two AUVs in Task II and III.

\subsection{Learning Curves}





Fig. \ref{learningcurve} shows the cumulative rewards received by the leader AUV and follower AUV during training via MAGAISIL with 
\textcolor{black}{sub-optimal} demonstrations, MAGAIL with sub-optimal and optimal demonstrations in Task I, measured by the predefined reward functions for leader and follower AUV in Section \ref{em}, respectively. From Fig. \ref{learningcurve} we can see that, for both leader and follower AUVs, at the beginning process, the speed of our MAGAISIL agent learning from sub-optimal demonstrations is similar to the MAGAIL agent learning from sub-optimal demonstrations, which is slower than the MAGAIL agent learning from optimal demonstrations. This might be because the good agent-generated trajectories selected by the human trainer in our MAGAISIL method did not fully replace the sub-optimal expert demonstrations yet. However, after about 400 episodes' training, our MAGAISIL agent learning from sub-optimal demonstrations reached a performance close to optimal expert demonstrations together with the MAGAIL agent learning from optimal demonstrations and statblized afterwards. In contrary, the performance of the MAGAIL agent learning from sub-optimal demonstrations still fluctuated around the sub-optimal demonstrations throughout the training process.

In summary, while the performance of MAGAIL agent learning from sub-optimal demonstrations is limited by sub-optimal expert demonstrations, our MAGAISIL agent can learn to reach a performance close to optimal demonstrations via gradually replacing the sub-optimal demonstrations with self-generated good trajectories selected by a human trainer.

\begin{figure*}[!t]
	\centering
	\subfloat[Task I]{
		\includegraphics[width=5.8cm]{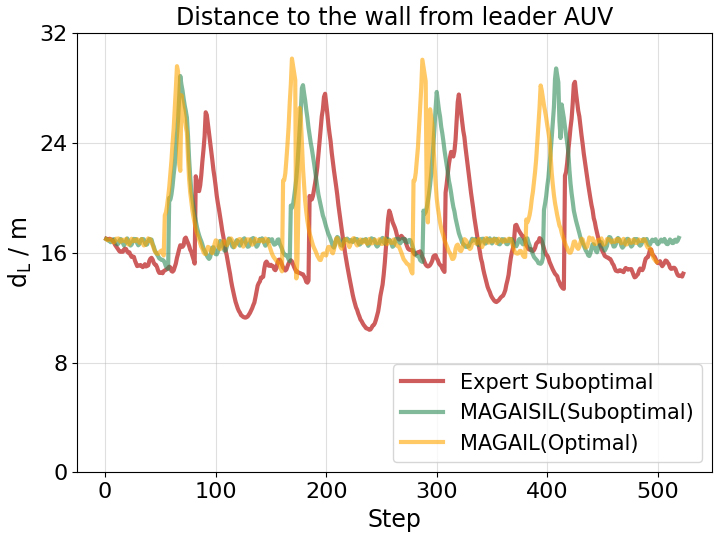}
	}
	\subfloat[Task II]{
		\includegraphics[width=5.8cm]{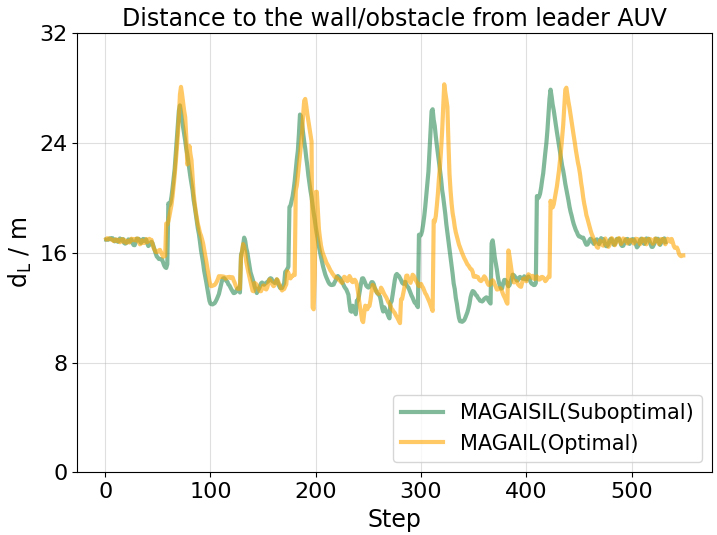}
	}
	\subfloat[Task III]{
		\includegraphics[width=5.8cm]{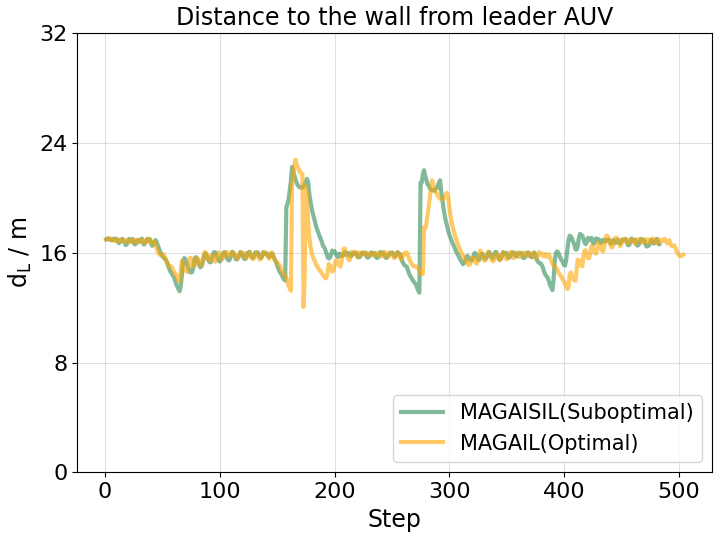}
	}
        \vspace{-3mm}
        \caption{\textcolor{black}{The distance to the wall of pipe or obstacles} from the leader AUV in Task I, II and III tested using final control policies trained in Task I via MAGAISIL with sub-optimal expert demonstrations and MAGAIL with optimal expert demonstrations. The red line in (a) shows \textcolor{black}{the distance to the wall of pipe} from the leader AUV during expert sub-optimal demonstration.} 
        \label{distanceWallLeader}
        \vspace{-5mm}
\end{figure*}

\begin{figure*}[!t]
	\centering
	\subfloat[Task I]{
		\includegraphics[width=5.8cm]{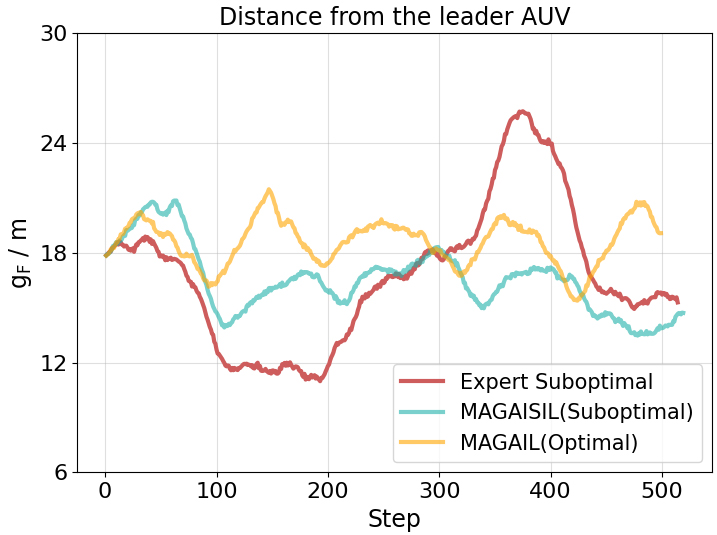}
	}
	\subfloat[Task II]{
		\includegraphics[width=5.8cm]{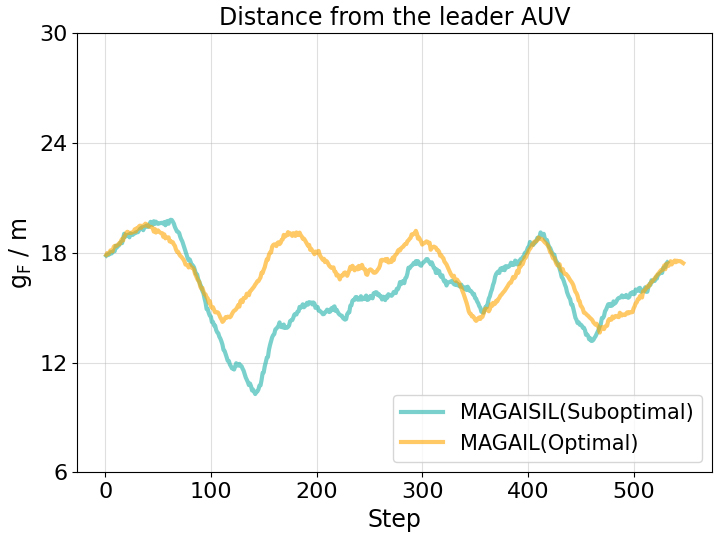}
	}
	\subfloat[Task III]{
		\includegraphics[width=5.8cm]{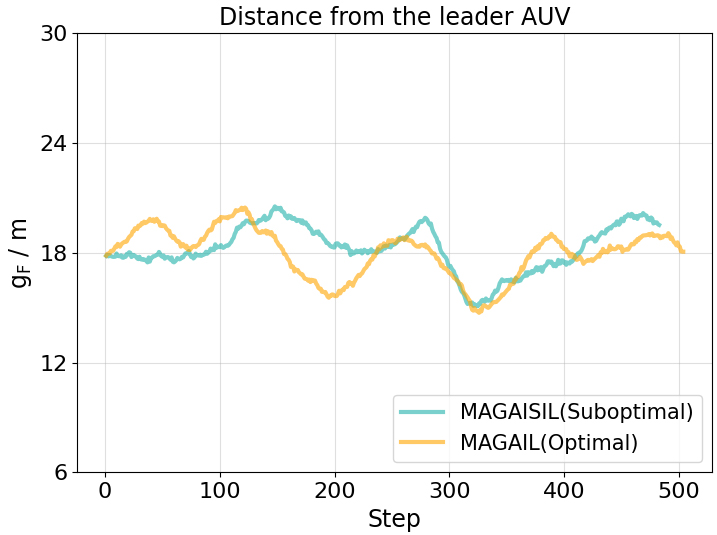}
	}
        \vspace{-3mm}
        \caption{The distance between the leader AUV and follower AUV in Task I, II and III tested using final control policies trained in Task I via MAGAISIL with sub-optimal expert demonstrations and MAGAIL with optimal expert demonstrations. The red line in (a) shows distance between the leader AUV and follower AUV during expert sub-optimal demonstration.}
        \label{distance-between}
        \vspace{-6mm}
\end{figure*}

\begin{figure*}[!t]
	\centering
	\subfloat[Task I]{
		\includegraphics[width=5.8cm]{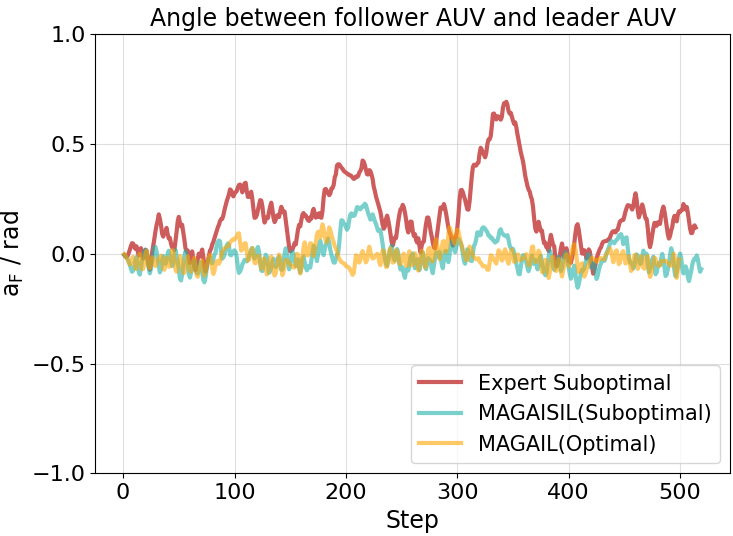}
	}
	\subfloat[Task II]{
		\includegraphics[width=5.8cm]{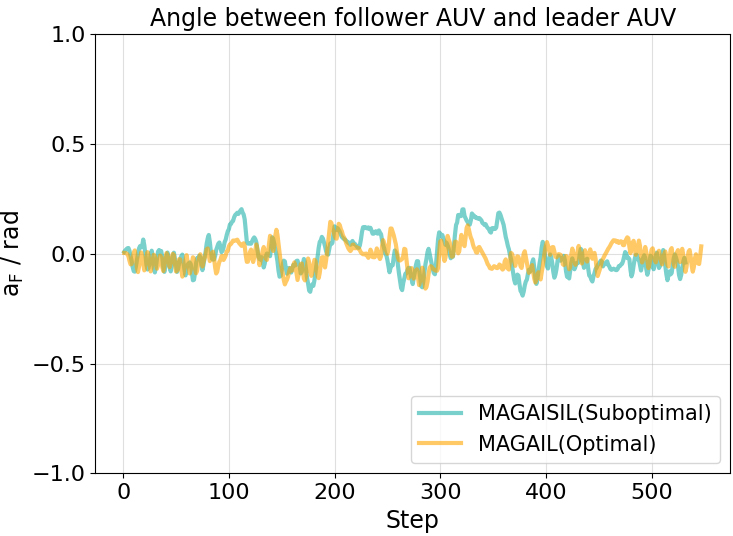}
	}
	\subfloat[Task III]{
		\includegraphics[width=5.8cm]{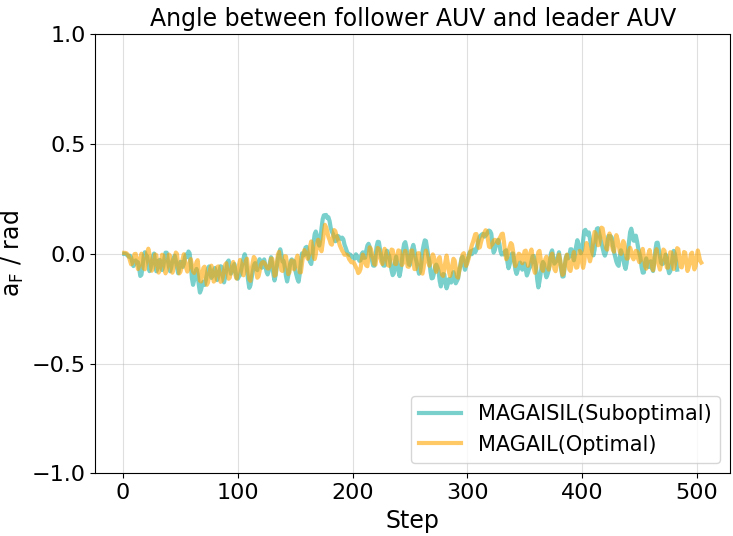}
	}
        \vspace{-3mm}
        \caption{The heading deviation of the follower AUV in Task I, II and III tested using final control policies trained in Task I via MAGAISIL with sub-optimal expert demonstrations and MAGAIL with optimal expert demonstrations. The red line in (a) shows heading deviation of the follower AUV during expert sub-optimal demonstration.}
        \label{heading-auv2}
        \vspace{-6mm}
\end{figure*}	
\begin{figure*}[!t]
	\centering
	\subfloat[Task I]{
		\includegraphics[width=5.78cm]{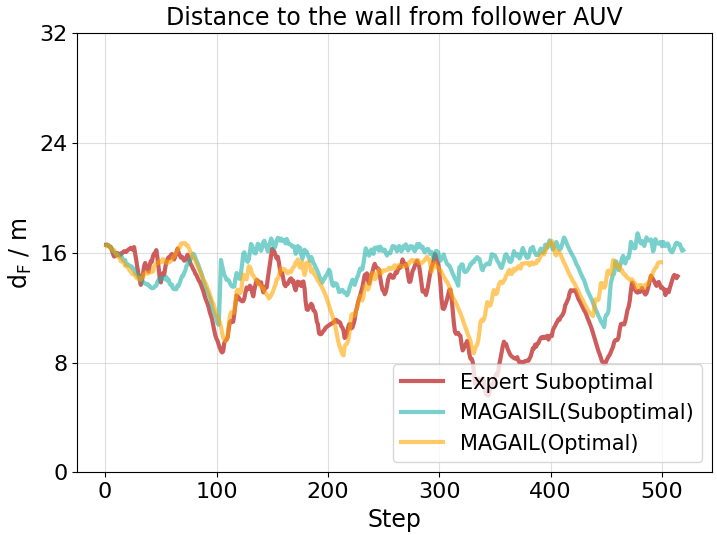}
	}
	\subfloat[Task II]{
		\includegraphics[width=5.78cm]{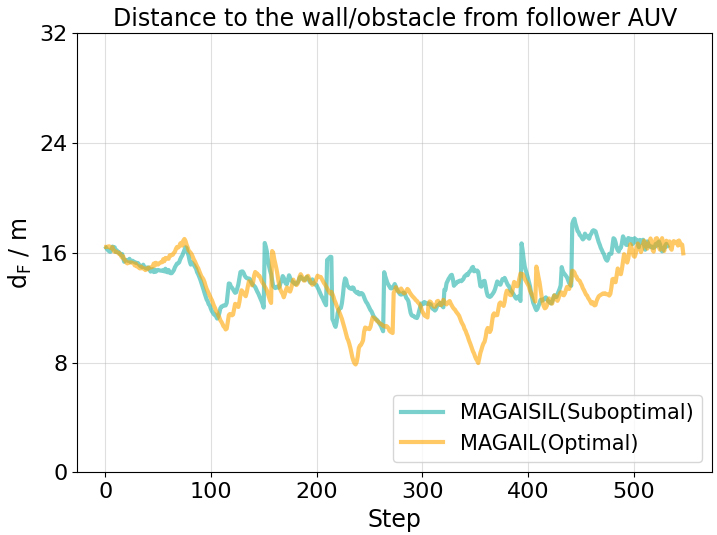}
	}
	\subfloat[Task III]{
		\includegraphics[width=5.78cm]{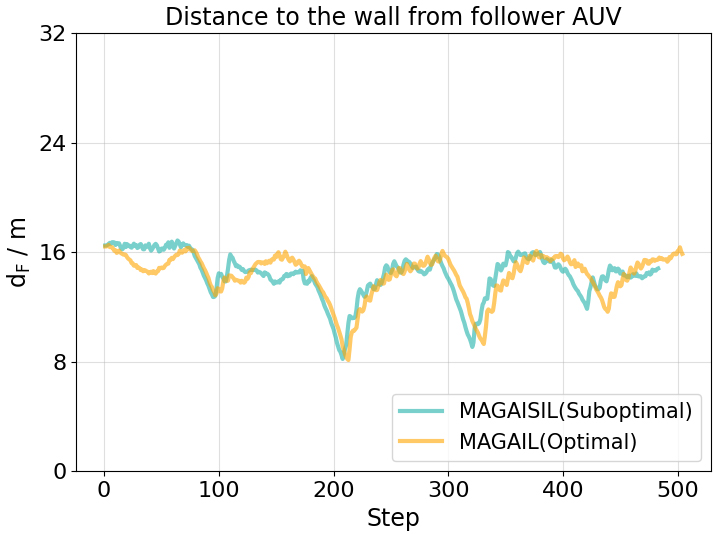}
	}
        \vspace{-3mm}
        \caption{\textcolor{black}{The distance to the wall of pipe or obstacles} from the follower AUV in Task I, II and III tested using final control policies trained in Task I via MAGAISIL with sub-optimal expert demonstrations and MAGAIL with optimal expert demonstrations. The red line in (a) shows \textcolor{black}{the distance to the wall of pipe} from the follower AUV during expert sub-optimal demonstration.}
        \label{distanceWall-auv2}
        \vspace{-6mm}
\end{figure*}	

\subsection{Performance}

We also tested and compared the final control policies of both leader and follower AUVs trained in Task I via our MAGAISIL with sub-optimal expert demonstrations and MAGAIL with optimal expert demonstrations from the perspectives of trajectory, distance to the walls of pipe, distance between leader and follower AUV, heading deviation of follower AUV, as shown in Fig. \ref{trajectory}(a), \ref{distanceWallLeader}(a), \ref{distance-between}(a), \ref{heading-auv2}(a), \ref{distanceWall-auv2}(a). From Fig. \ref{trajectory}(a) we can see that, both leader and follower AUV trained via our MAGAISIL with sub-optimal demonstrations and MAGAIL with optimal demonstrations can successfully complete Task I by generally following the middle of the pipe. The leader and follower AUV trained via our MAGAISIL method even performed a bit better than the one via MAGAIL at the turnings in the pipe. 
While further examining the observations of both leader and follower AUV in the testing process, we found that although the leader AUV in the provided sub-optimal expert demonstrations fluctuated dramatically around the middle of the pipe after turning, the leader AUV trained with our MAGAISIL method can get a performance close to MAGAIL learning from optimal demonstrations, which can immediately get back to the safe distance around 17.3 meters (derived based on the detection angle range of the sonar sensor and the distance between walls of the pipe, refer to Section \ref{em}), and keep itself in the middle of the pipe (Fig. \ref{distanceWallLeader}(a)).

For the follower AUV in the provided sub-optimal expert demonstrations, its distance to the leader AUV fluctuated dramatically between 12 and 25 meters and heading deviation also fluctuated dramatically (Fig. \ref{distance-between}(a) and \ref{heading-auv2}(a)). After training with our MAGAISIL method, the follower AUV can obtain a performance close to MAGAIL learning from optimal demonstrations, and keep a distance to the leader AUV at around 18 meters with a heading deviation around 0 (Fig. \ref{distance-between}(a) and \ref{heading-auv2}(a)). Moreover, the follower AUV trained with our MAGAISIL method can keep a safe distance to the walls of pipe even though it fluctuated largely during and after turnings in the provided sub-optimal expert demonstrations, and even performed better than the MAGAIL agent learning from optimal demonstrations (Fig. \ref{distanceWall-auv2}(a)). 

In summary, our results show that 
both leader and follower AUV trained with our MAGAISIL method can surpass the provided sub-optimal expert demonstrations and get a performance close to or even better than those trained via MAGAIL with optimal demonstrations.

\subsection{Adaptability to Complex and Different Tasks}

We also tested and compared the adaptability of our MAGAISIL method to complex and different tasks by running the saved final control policies of leader and follower AUV trained in Task I via MAGAISIL with sub-optimal demonstrations and via MAGAIL with optimal demonstrations in Task II with extra obstacles and in Task III with changed angles of the walls and extended length of the pipe. Other settings are the same as Task I. 

Fig. \ref{trajectory}(b)(c), \ref{distanceWallLeader}(b)(c), \ref{distance-between}(b)(c), \ref{heading-auv2}(b)(c), \ref{distanceWall-auv2}(b)(c) show the performance of the leader and follower AUV trained via MAGAISIL and MAGAIL in terms of trajectory, distance to the walls of pipe, distance between leader and follower AUV, heading deviation of follower AUV, respectively. From these results in Task II and III we can see that, the control policies of leader and follower AUV trained via MAGAISIL and MAGAIL can adapt well to complex and different tasks even with added obstacles or changed angles of wall. Moreover, the both AUVs trained via MAGAISIL with sub-optimal demonstrations can obtain a similar performance to those trained via MAGAIL with optimal demonstrations. The leader and follower AUV trained via MAGAISIL even performed a bit better than those via MAGAIL at the turnings in the pipe (Fig. \ref{trajectory}(b)(c)). However, the distance between follower AUV and leader AUV trained via MAGAISIL decreased largely after 100 steps compared to those via MAGAIL, but gradually increased after that and is similar to those trained via MAGAIL after about 300 steps (Fig. \ref{distance-between}(b)). This might be because of the effect of added obstacles, which were first met after the first turning at about 100 steps. This is consistent with the a bit larger fluctuation of the heading deviation of the follower AUV trained via MAGAISIL compared to MAGAIL, which also starts from about 100 steps (Fig. \ref{heading-auv2}(b)).

\section{Conclusion}


In this paper, we builds upon the MAGAIL algorithm by proposing multi-agent generative adversarial interactive self-imitation learning (MAGAISIL), which can facilitate AUVs to learn policies by gradually replacing the provided sub-optimal demonstrations with self-generated good trajectories
selected by a human trainer. We tested and implemented MAGAISIL in a multi-AUV
formation control and obstacle avoidance task on the Gazebo platform with AUV simulator of our lab and compared with MAGAIL. Results show that, AUVs trained with our MAGAISIL method can surpass the provided sub-optimal expert demonstrations and learn to reach a performance close to or even better than those trained via MAGAIL with optimal demonstration. Further results indicate that the control policies of AUVs trained via MAGAISIL can adapt to complex and different tasks as well as MAGAIL learning from optimal demonstrations. 




\section*{ACKNOWLEDGMENT}

This work was supported by Natural Science Foundation of China (under grant No. 51809246) and Qingdao Municipal Natural Science Foundation (under grant No. 23-2-1-153-zyyd-jch).


\bibliographystyle{IEEEtran}
\bibliography{ref}

\end{document}